\documentclass[10pt,twocolumn,letterpaper]{article}

\usepackage[pagenumbers]{wacv} 
\definecolor{wacvblue}{rgb}{0.21,0.49,0.74}
\usepackage[pagebackref,breaklinks,colorlinks,allcolors=wacvblue]{hyperref}
\usepackage{import}
\usepackage{dsfont}
\usepackage{url}
\usepackage{multirow} 
\usepackage{graphicx}
\usepackage{amsmath}
\usepackage{amsthm}
\usepackage{amssymb}
\usepackage{booktabs}
\usepackage{ulem}
\usepackage[T1]{fontenc}

\title{ Prithvi-Complimentary Adaptive Fusion Encoder (CAFE): unlocking full-potential for flood inundation mapping }

\author{
    Saurabh Kaushik$^{1,*}$ \quad
    Lalit Maurya$^{2,*}$\quad
    Beth Tellman$^{1}$\\[0.2cm]
    $^{1}$Center for Sustainability and the Global Environment (SAGE), University of Wisconsin–Madison, \\Madison, WI, 53726 USA\\
    $^{2}$Portsmouth AI and Data Science Centre (PAIDS), School of Computing, University of Portsmouth, \\Portsmouth, PO1 3HE, UK \\
    {\tt\small skaushik8@wisc.edu,
    lalit.maurya@port.ac.uk,
    beth.tellman@wisc.edu}
}

\begin{document}
\maketitle
\let\thefootnote\relax\footnotetext{${}^{*}$These authors contributed equally.}
\begin{abstract}
Geo-Foundation Models (GFMs), have proven effective in diverse downstream applications, including semantic segmentation, classification, and regression tasks. However, in case of flood mapping using Sen1Flood11 dataset as a downstream task, GFMs struggles to outperform the baseline U-Net, highlighting model's limitation in capturing critical local nuances. To address this, we present the Prithvi-Complementary Adaptive Fusion Encoder (CAFE), which integrate Prithvi GFM pretrained encoder with a parallel CNN residual branch enhanced by Convolutional Attention Modules (CAM). Prithvi-CAFE enables fast and efficient fine-tuning through adapters in Prithvi and performs multi-scale, multi-level fusion with CNN features, capturing critical local details while preserving long-range dependencies. We achieve state-of-the-art results on two comprehensive flood mapping datasets: Sen1Flood11 and FloodPlanet. On Sen1Flood11 test data, Prithvi-CAFE (IoU 83.41) outperforms the original Prithvi (IoU 82.50) and other major GFMs (TerraMind 82.90, DOFA 81.54, spectralGPT: 81.02). The improvement is even more pronounced on the hold-out test site, where Prithvi-CAFE achieves an IoU of 81.37 compared to the baseline U-Net (70.57) and original Prithvi (72.42). On FloodPlanet, Prithvi-CAFE also surpasses the baseline U-Net and other GFMs, achieving an IoU of 64.70 compared to U-Net (60.14), Terramind (62.33), DOFA (59.15) and Prithvi 2.0 (61.91). Our proposed simple yet effective Prithvi-CAFE demonstrates strong potential for improving segmentation tasks where multi-channel and multi-modal data provide complementary information and local details are critical. The code is released on \href{https://github.com/Sk-2103/Prithvi-CAFE}{Prithvi-CAFE Github} 

\end{abstract}
    
\section{Introduction}
\label{sec:intro}
Geo-Foundation Models (GFMs) are built on self-supervision techniques, primarily Masked Autoencoders (MAE) \cite{he2022masked} \cite{szwarcman2024prithvi}\cite{xiong2024neural} and contrastive learning \cite{tseng2025galileo} \cite{guo2024skysense}, to leverage the abundance of unlabeled remote sensing data emerging from diverse sensors \cite{lu2025vision}. GFMs present an alternative approach from longstanding  application specific model to task-agnostic models that can produce reliable maps with sparse labels, owing to GFMs’ massively pretrained encoders \cite{szwarcman2024prithvi}\cite{xiong2024neural}.  In this quest, more than 50 large vision foundation models have been proposed \cite{lu2025vision}, showing potential in domains such as wildfire, marine, agriculture, urban land cover, forest, and floods using community benchmark datasets, GeoBench \cite{lacoste2023geo} and PANGEA \cite{marsocci2024pangaea}. However, interesting observations arise in several domains including flood inundation mapping using Sen1Floods11 data \cite{bonafilia2020sen1floods11}: current GFMs under-perform the baseline U-Net so far. This becomes even more intriguing given detailed experiments and comparisons by \cite{marsocci2024pangaea}, highlighting the U-Net \cite{ronneberger2015u} consistently outperforms all GFMs even in limited-data scenarios (e.g., 10\%, 50\%, and 100\% of training data). These results are critical for Earth scientists and the machine learning community regarding the utility of such large models with hundreds of millions of parameters demand high computational resources yet still under-perform compared to the baseline U-Net, a much lighter model with only 31M parameters.

These observations highlight the limitations of large pre-trained GFMs in efficiently learning critical local representations compared to much simpler and lighter U-Net models. They underscore ample scope to advance GFM architectures in terms of capturing essential local nuances and supporting any number of input channels, especially in the case of the Prithvi GFM \cite{szwarcman2024prithvi}, which currently allows only six input channels. This restriction stems from its original pre-training on six Harmonized Landsat and Sentinel-2 (HLS) spectral bands using 4.2M training samples. Such an architectural design limits its ability to generalize across many Earth observation applications that leverage multispectral and multi-modal data. In this pursuit, we opted for the Prithvi GFM pretrained encoder, given its popularity in diverse Earth observation applications \cite{szwarcman2024prithvi} and its adaptation beyond community-standard datasets like GeoBench \cite{lacoste2023geo} and PANGEA \cite{marsocci2024pangaea} for example, in glacial lake mapping \cite{jiang2025glacial} and debris-covered glacier mapping \cite{kaushik2025debris}. 

To address these limitations, we propose the Prithvi-Complementary Adaptive Fusion Encoder (CAFE), which can handle any number of channels. Our approach leverages the strengths of both Transformers and CNNs through complementary channel fusion. We pass six channels (on which Prithvi was pretrained originally) through the Prithvi encoder, and use an adapter-based fast and efficient fine-tuning method, reducing trainable parameters from 650M to 45.5M.  All other channels are processed through a CNN residual block and a Convolutional Attention Module (CAM). 
To Integrates Transformer and CNN features, we adopt a multiscale and multilevel attention-based fusion approach that preserves the most relevant gobal and long-range information. Mathematically,:
\[
\text{Transformer: } f_1(x_{S_1}), \quad 
\text{CNN: } f_2(x_{S_2}), \quad 
S_1 \cap S_2 = \emptyset
\]
The fusion module learns a joint embedding:
\[
F = f_{\text{fuse}}\big(f_1(x_{S_1}), f_2(x_{S_2})\big)
   \approx f^{*}(x_{S_1 \cup S_2})
\]
i.e., approximating the full-spectrum feature map through learned cooperation. Our contribution are:
\begin{itemize}
    \item We propose simple yet effective segmentation models that maximize critical local nuances while preserving long-range dependencies.
    \item We extend Prithvi-GFM’s capability beyond six spectral channels to efficiently process any number of channels and produce reliable flood maps.  
    \item We report state-of-the-art (SoTA) results on the benchmark flood dataset (i.e., Sen1Flood11) through efficient fine-tuning and effective multi-scale, multi-level attention fusion.
 \end{itemize}

\section{Related Work}
\subsection{Geo-Foundational Models}
GFMs remain the focus of a large scientific community aiming to address long-standing limitations of task-specific models, 1) under-utilization of unlabeled remote sensing datasets, 2) exclusion of temporal components in Earth observation, 3) generating reliable maps with sparse labels and 4) underexplored capabilities of multimodal data. As a result, we have witnessed the development of 58 remote sensing vision foundation models \cite{lu2025vision}, offering unprecedented opportunities to push the boundaries of Earth observation. Unlike application-specific models, GFMs have proven efficient in various downstream tasks, including classification, semantic segmentation, change detection, and regression across diverse domains\cite{szwarcman2024prithvi}\cite{xiong2024neural}\cite{clay2024foundation} \cite{tseng2025galileo}\cite{hong2023spectralgpt}\cite{jakubik2025terramind}\cite{wang2024decoupling} \cite{hong2023spectralgpt}. To assess the performance of these large vision models, the community has curated two primary benchmark datasets GeoBench \cite{lacoste2023geo} and PANGEA \cite{marsocci2024pangaea}, to evaluate GFM performance across various domains such as forestry, floods, and land use/land cover.

A discussion of all 58 foundation models is beyond the scope of this article; therefore, we highlight some popular and recent models. Recently, Google released its AlphaEarth Foundation (AEF) model \cite{brown2025alphaearth}, the first of its kind to employ time-continuous embeddings by training on 8,412,511 video sequences collected from nine publicly available satellite sensors, high-quality land cover maps, and text, rather than treating satellite imagery as a single timestamp. However, unlike other foundation models such as Clay \cite{clay2024foundation}, Prithvi \cite{szwarcman2024prithvi}, DOFA \cite{xiong2024neural}, Panopticon \cite{waldmann2025panopticon}, and Galileo \cite{tseng2025galileo}, the AlphaEarth model is not open-source, limiting its applicability for fine-tuning on specific Earth observation tasks. In contrast, many fully open-access GFMs can be fine-tuned for any Earth observation task, offering greater flexibility for monitoring short-lived changes or specific events. Among these, Clay GFM \cite{clay2024foundation} is one of the largest, pretrained on ~70 million image chips from seven diverse remote sensing sensors. Clay is based on an MAE ViT backbone, similar to Prithvi \cite{szwarcman2024prithvi}, DOFA \cite{xiong2024neural}, ScaleMAE \cite{reed2023scale}, and Spectral-GPT \cite{hong2023spectralgpt}. Among these models, Prithvi 2.0 shows promising results on the GeoBench dataset, surpassing other models (DOFA ViT-300, DINO \cite{wang2023ssl4eo}, Decur \cite{wang2024decoupling}, Satlas \cite{bastani2023satlaspretrain}, ScaleMAE \cite{reed2023scale}) in overall performance. However, while models like DOFA and Clay allow any number of input channel.

Prithvi’s original architecture only supports the six spectral bands on which it was pretrained, limiting its applicability for Earth observation tasks requiring multimodal input data. Nevertheless, given its potential and ease of accessibility via TerraTorch \cite{gomes2025terratorch}, Prithvi remains one of the most popular models adapted for various Earth observation applications beyond community benchmarks (e.g., \cite{kaushik2025debris}\cite{jiang2025glacial}). This motivates the present work to extend Prithvi’s capabilities through efficient fusion with CNN blocks, enabling support for any input channel and improved understanding of critical local information. Recently, generative models such as TerraMind \cite{jakubik2025terramind}, and DiffusionSat \cite{khanna2024diffusionsat}  have also been proposed, aiming to generate artificial data to fill gaps and improve performance using synthetically generated complementary data. Vision-language GFMs (VLGFMs) such as RemoteCLIP \cite{liu2024remoteclip}, RSGPT \cite{hu2025rsgpt}, and SkySenseGPT \cite{zhan2025skyeyegpt} have also emerged, pushing the boundaries of Earth observation by enabling users to interact with geospatial data through natural language prompts.

\subsection{Geo-Foundation Models beyond benchmark dataset}

GFMs have been extensively assessed on benchmark datasets \cite{marsocci2024pangaea} \cite{lacoste2023geo}; however, evaluation of GFMs outside benchmarks also bears great significance as it directly highlights the models’ performance on new domains, datasets, and extensive comparisons with state-of-the-art models. In this quest, Prithvi remains one of the most widely used models for various downstream tasks, including land cover mapping \cite{lambhate2024finetuning}, demonstrated fine-tuning Prithvi GFM for land cover mapping and achieved an mIoU of 62.37 compared to baseline U-Net (36.36) and ViT (46.8). Prithvi GFM also found to effecient in glacial lake mapping. UViT \cite{jiang2025glacial} a U-Net-style Vision Transformer (ViT) leveraging the Prithvi pretrained encoder and enhanced squeeze-and-excitation layers to incorporate multi-sensor data. A similar study by \cite{kaushik2025debris} reported an 8\% mIoU improvement in mapping debris-covered glaciers at global scale compared to U-Net. The successful implementation of Prithvi in a completely different domain (Cryosphere) which is underrepresented in benchmark datasets, exhibits the model’s versatility for downstream tasks. Kostejn et al. \cite{kostejn2025u} proposed the U-Prithvi model for flood segmentation, combining the strength of the pretrained Prithvi encoder with U-Net and evaluating the model’s performance on the Sen1Floods11 dataset. In U-Prithvi, input is passed through both U-Net and Prithvi encoder and subsequently fed to the decoder with skip connections. The model shows ~5\% improvement on geographically held-out test sites.

\label{related_work}
\section{Method}
\label{sec:method}
\begin{figure*}
    \centering
    \includegraphics[width=1\linewidth]{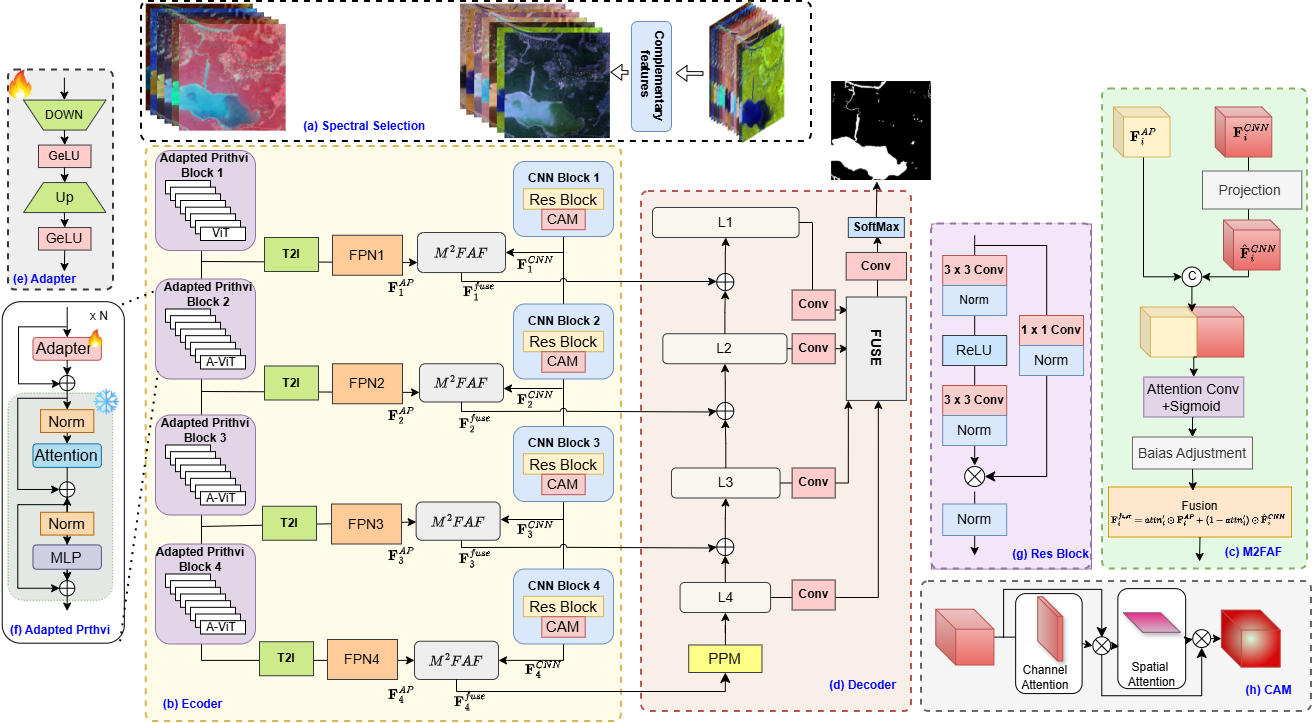}
    \caption{The proposed Prithvi-CAFE architecture. (a) The Spectral Selection module divides input images into two spectral branches. (b) The encoder processes these branches using adapted Prithvi blocks and CNN blocks, respectively. (c) The Multi-Scale Multi-Level Feature Attention Fusion (M$^{2}$FAF) module merges features from both streams. (d) The Decoder integrates them via pyramid pooling and lateral connections to generate the segmentation mask. (e) Adapter modules are attached to each ViT block of Prithvi, as shown in (f). (g) Residual Blocks and (h) Convolutional Attention Modules (CAM) are used to enhance CNN features. }
    \label{fig:block}
\end{figure*}
Here, we explained the architecture of our proposed Prithvi-CAFE as shown in the Fig \ref{fig:block}. The Prithvi-CAFE architecture begins by dividing the input image tensor into two complementary parts: one directed to the Adapted Prithvi transformer and the other to the CNN branch. Selected spectral bands are allocated to the transformer to capture rich spectral features, while the remaining channels are processed by the CNN to extract detailed spatial information (Fig. \ref{fig:block} (a)). The Adapted Prithvi transformer incorporates per-block dynamic prompt adapters (Fig. \ref{fig:block} (e)), which enable efficient fine-tuning of a frozen pretrained model using lightweight residual modules, significantly reducing computational cost (Fig. \ref{fig:block} (f)). Outputs from multiple transformer layers are reshaped from token sequences into two-dimensional feature maps representing different semantic levels. These are refined through Feature Pyramid Networks that upsample and adjust channel dimensions to create hierarchical multi-scale representations (Fig. \ref{fig:block} (b)). In parallel, the CNN backbone processes its input through several Residual Blocks integrated with Convolutional Attention Modules, which enhance important feature channels and spatial regions. The resulting CNN feature maps are resized to match the transformer outputs, enabling effective fusion. The Multi-Scale Multi-Level Feature Attention Fusion module then adaptively merges contextual transformer features with fine spatial cues from the CNN using attention weighting (Fig. \ref{fig:block} (c)). Finally, the UperNet decoder integrates the fused features through pyramid pooling and lateral connections to generate precise and context-aware segmentation outputs (Fig. \ref{fig:block} (d)).
\subsection{Complementary Feature}
Let the input tensor be denoted as:$X \in \mathbb{R}^{B \times C \times H \times W} $, where $B$ is the batch size, \(C\) the number of spectral channels, and \(H, W\) the spatial dimensions. To leverage the pretrained Prithvi transformer effectively, the input channels are split into two complementary subsets:
\begin{equation}
    X = [X_{\text{AP}} \,|\, X_{\text{CNN}}]
\end{equation}
where \(X_{\text{AP}}\) is the subset of spectral channels fed to the Adapted Prithvi encoder, and \(X_{\text{CNN}}\) contains the remaining channels for the CNN pathway. Formally if,
\begin{equation}
\small
    \mathcal{I}_{\text{AP}} = \{1, 2, 3, 7, 11, 12\}, \quad 
\mathcal{I}_{\text{CNN}} = \{1, \dots, C\} \setminus \mathcal{I}_{\text{AP}}
\end{equation}

then the input partitions become:
\begin{equation}
\small
    X_{\text{AP}} = X[:, \mathcal{I}_{\text{AP}}, :, :], \qquad
X_{\text{CNN}} = X[:, \mathcal{I}_{\text{CNN}}, :, :].
\end{equation}
This complementary feature separation ensures that the transformer processes spectrally-rich features, while the CNN branch captures fine-grained spatial patterns. By routing other informative channels to the CNN, the model digs rich contents and efficiently leverages Prithvi’s pretrained representations, which is particularly beneficial for multi-spectral data where some bands contribute differently with regards to local and global context, and balances spectral vs. spatial learning.

\subsection{Parallel Backbones: Transformer + CNN Pathways}

\subsubsection{Transformer Backbone - Adapted Prithvi}
The transformer backbone employs a Dynamic Prompt Adapter for parameter-efficient fine-tuning of Prithvi ViT blocks. In this design, a lightweight learnable adapter is positioned at the beginning of each ViT (as shown in Fig. \ref{fig:block} (f)), dynamically adjusting prompts to guide task-specific adaptation while keeping the pretrained Prithvi encoder frozen. This arrangement reduces memory and computational cost, and enables efficient transfer to new datasets without full retraining, while preserving the model’s ability to capture long-range dependencies. Formally, let $\mathrm{blk}(\cdot)$ denote a Prithvi transformer block. The adapted block is defined as:
\begin{equation}
y = \mathrm{blk}(x + f_{\text{adapter}}(x))
\end{equation}
\begin{equation}
f_{\text{adapter}}(x) = \sigma \left( W_2 \cdot \sigma(W_1 \cdot x + b_1) + b_2 \right)
\end{equation}
where, $W_1 \in \mathbb{R}^{d \times 32}$, and $\quad W_2 \in \mathbb{R}^{32 \times d}$. The adapter adds a low-rank residual perturbation $f_{\text{adapter}}(x)$ (bottlenecked through 32 dimensions for efficiency), 
enabling the model to adapt representations without retraining the billions of parameters in Prithvi (600M parameters). Adapters thus provide a low-rank correction, preserving pretrained distributions while tailoring representations to new datasets. Fine-tuning with adapters reduces GPU memory usage by 50–80\% compared to full unfreezing.

From the transformer outputs, features are extracted at layers $\{7, 15, 23, 31\}$ correspond to multi-level semantic depths. These token maps $f_i^{AP} \in \mathbb{R}^{B \times N_i \times C_i}$ are reshaped back into 2D spatial tensors using a token-to-image (T2I) transformation, where $C_i$ is tokken embedding which is 1280 for Prithvi.
\begin{equation}
    f_i^{AP} \in \mathbb{R}^{B \times C_i \times H_i \times W_i}, \quad i \in \{7, 15, 23, 31\}.
\end{equation}
Each feature map is further refined via Feature Pyramid Networks (FPN) to align spatial resolutions and reduce channel dimensions for multi-scale fusion:
\begin{equation}
    \mathbf{F}_i^{AP} = \text{FPN}_i(f_i^{AP})
\end{equation}
The FPN modules apply convolutional upsampling to the selected transformer features to align spatial resolutions and progressively reduce channel dimensions for effective multi-scale fusion. Specifically, FPN1 performs an $8\times$ upsampling using three ConvTranspose2d layers, reducing channels from 1280 → 640 → 320 → 160. FPN2 performs a $4\times$ upsampling, reducing channels from 1280 → 640 → 320. FPN3 applies a $2\times$ upsampling, reducing channels from 1280 → 640, while FPN4 maintains the spatial size using a $1\times1$ convolution with 1280 channels. These hierarchical refinements enhance spatial detail recovery and ensure consistent feature alignment for downstream fusion and segmentation tasks. This process produces a set of multi scale multi level transformer features: 
\begin{equation}
    \mathbf{F}^{AP} = \{\mathbf{F}_1^{AP}, \mathbf{F}_2^{AP}, \mathbf{F}_3^{AP}, \mathbf{F}_4^{AP}\}
\end{equation}

\subsubsection{CNN Backbone}
The CNN Backbone processes $X_{\text{CNN}}$
through four Residual Blocks \cite{he2016deep} with CAM. The Residual Block enhances feature learning through residual connections. Each residual block learns:
\begin{equation}
    y = \sigma(x) + F(x; W)
\end{equation}
 where $F(x; W)$ consists of two $3\times3$ convolutional layers followed by batch normalization and ReLU activation, along with a shortcut path $\sigma(.)$ that  performs a $1\times1$ convolution when input and output dimensions differ. This design allows the block to learn residual mappings, improving gradient flow and network stability.
The CAM (Fig. \ref{fig:block} (h)) improves CNN feature representations by applying channel and spatial attention sequentially. Given an input feature F, channel attention generates weights using global average and max pooling, emphasizing important feature channels. Spatial attention then highlights important regions by combining pooled spatial maps and applying a convolution. This dual attention mechanism adaptively focuses on what and where to attend in the feature map, enhancing feature discrimination, representation power, and overall performance in segmentation task. By reducing noise (e.g., atmospheric artifacts), CAM improves feature quality with minimal parameter overhead. The resulting process produces a set of hierarchical CNN features: 
\begin{equation}
   \mathbf{F}^{CNN} =  \{\mathbf{F}_1^{CNN}, \mathbf{F}_2^{CNN}, \mathbf{F}_3^{CNN}, \mathbf{F}_4^{CNN}\}
\end{equation}

\subsection{Multi-Scale Multi level Feature Attention Fusion (M$^{2}$FAF)}
An attention-driven fusion block designed to unify contextual and fine-grained features across scales and levels. Adapted Prithvi features (after FPN) are fused with CNN features using attention-based weighting. After feature extraction, both backbones yield four-scale feature maps, Each CNN feature map is upsampled and projected to match the corresponding transformer feature resolution:
\begin{equation}
    \hat{\mathbf{F}}_i^{CNN} = \text{Conv}_{1\times1}\big(\text{Interpolate}({\mathbf{F}}_i^{CNN}, \text{size}({\mathbf{F}}_i^{AP}))\big)
\end{equation}
An attention mask is then computed for each scale:
\begin{equation}
   \mathbf{A}_i = \sigma(\text{Conv}_{1\times1}([{\mathbf{F}}_i^{AP}; {\mathbf{F}}_i^{CNN}])) 
\end{equation}
where \(\sigma\) denotes the sigmoid activation and \([\cdot ; \cdot]\) indicates channel concatenation. To stabilize the fusion, a  bias factor \(\beta\) is applied:
\begin{equation}
    {attn}_i' = {attn}_i (1 - \beta) + \beta, \quad \text{with } \beta \in [0,1].
\end{equation}
The final multi-scale fused representation is obtained as:
\begin{equation}
    \mathbf{F}_i^{fuse} = attn_i' \odot \mathbf{F}_i^{AP} + (1 - attn_i') \odot \hat{\mathbf{F}}_i^{CNN}, \quad i = 1, \dots, 4.
\end{equation}
Thus, each fused feature \({\mathbf{F}}_i^{fuse}\) adaptively combines contextual transformer information with spatially-dense CNN cues:
Here, \(attn_i' \approx 1\) favors transformer features (${\mathbf{F}}_i^{AP}$), while \(attn_i' \approx 0\) favors CNN features (\(\hat{\mathbf{F}}_i^{CNN}\)).

\subsection{UperNet Decoder}
The UperNet Decoder (Fig. \ref{fig:block} (d))combines multi-scale features from a backbone for segmentation mask genretaion. It processes four feature maps, from high-resolution detailed maps to low-resolution semantic maps. The Pyramid Pooling Module (PPM) captures global context by pooling the smallest feature at multiple scales and merging them. Lateral connections (L4-L1) reduce channel dimensions of each feature map. Features are then fused progressively from the smallest to the largest, integrating high-level semantic information with fine-grained spatial details. Finally, a convolutional layer refines the fused features to produce the final output, resulting in precise and context-aware predictions.

\section{Experiments}
\textbf{Dataset and Metrics.} We evaluated our proposed Prithvi-CAFE on two high-quality flood datasets: Sen1Floods11 \cite{bonafilia2020sen1floods11} and FloodPlanet \cite{zhang2025assessing}, given the limited performance of most GFMs in flood mapping, where U-Net consistently outperforms all GFMs. This makes flood mapping an ideal challenge to test the model’s capability for efficient fusion of long-range information captured by transformers and detailed spatial information extracted by CNNs. Sen1Floods11 consists of 446 hand-labeled image-labels pairs using Sentinel-1 and Sentinel-2 data collected worldwide. This data set is one of the widely used benchmark data sets to evaluate the performance of GFM for flood mapping as a downstream task, allowing direct comparison of our model’s performance with SoTA results. The second dataset, FloodPlanet, consists of 19 major flood events worldwide. This dataset represents one of the highest-quality datasets for evaluating deep learning models, since all labels are manually mapped using high-resolution PlanetScope imagery. These high-resolution labels significantly improved model performance (15.6), even when relying on moderate-resolution satellite imagery (10 m Sentinel-2). Thus, combining these two datasets for evaluation introduces sufficient complexity for flood segmentation tasks.

As an evaluation metric, we opted for standard segmentation metrics such as Intersection over Union (IoU) for the foreground class (flood in our case) mean Intersection over Union (mIoU).  IoU measures the overlap between the predicted segmentation and the ground truth, calculated as the ratio of their intersection to their union. We also computed mean Dice score, also known as the m-F1 score, which represents the harmonic mean of precision and recall, providing a balanced measure of accuracy for imbalanced classes.\\

\noindent\textbf{Implementation Details.} To implement Prithvi-CAFE on Sen1Floods11, we used the exact given train, test, and validation split given by \cite{bonafilia2020sen1floods11} to keep the results fairly comparable. FloodPlanet consists of 298 Sentinel-2 images paired with PlanetScope-derived labels. The original data set had inconsistent image sizes close to 320 px; therefore, we resized all images and labels to a consistent 320×320 before feeding them to the model. Since FloodPlanet does not have a predefined split, we performed a 4-fold cross-validation, where 70 of the data was used for training, 10 for validation and 20 for testing in each split. This approach provides a large number of heterogeneous examples for testing and reduces model's bias toward any particular use case. We used cross-entropy loss, the AdamW optimizer for model convergence, and a StepLR scheduler. To optimize hyperparameters such as learning rate, optimizer parameters (weight decay), and scheduler parameters (step size and gamma), we ran 20 trials, each with 60 epochs and patience of 10 for early stopping. The best hyperparameters obtained on validation data were used for full model training with a patience level of 20 for early stopping. To ensure comparability across experiments, we fixed the random seed to 42, used a batch size of 8 and conducted all experiments on an NVIDIA RTX A6000 GPU. 

\section{Results and Discussion}
\subsection{Model’s evaluation on Sen1Floods11}
Our proposed Prithvi-CAFE performed very well on the Sen1Flood11 \cite{bonafilia2020sen1floods11} test set, achieving 83.41 IoU water, outperforming most GFMs, including Prithvi-600M (82.50), Prithvi-300M (82.20), TerraMind (82.90), DOFA (81.54), and other recently proposed models including DeepSAR \cite{sharma2025deepsarflood} (72.22) and MM UNet \cite{portales2025understanding} (73.84) (Table \ref{flood_table}). U-Net shows only marginal improvement (0.63) compared to Prithvi-CAFE. In the case of a geographically held-out test site (i.e., Bolivia), Prithvi-CAFE surpasses the baseline U-Net by 10.8 IoU, and recently proposed U-Prithvi (which also leverages U-Net with a Prithvi encoder) by 1.69, and the original Prithvi by 9 (Table \ref{flood_table}). Prithvi-CAFE further outperforms adapter-based methods, achieving mIoU 88.87 compared to ViT Adapter (84.94) and LoRA Adapter (87.57) (Table \ref{flood_table}). These findings highlight that our proposed model demonstrates strong spatial transferability compared to other models. \begin{figure}[!h]
    \centering
    \includegraphics[width=0.7\linewidth]{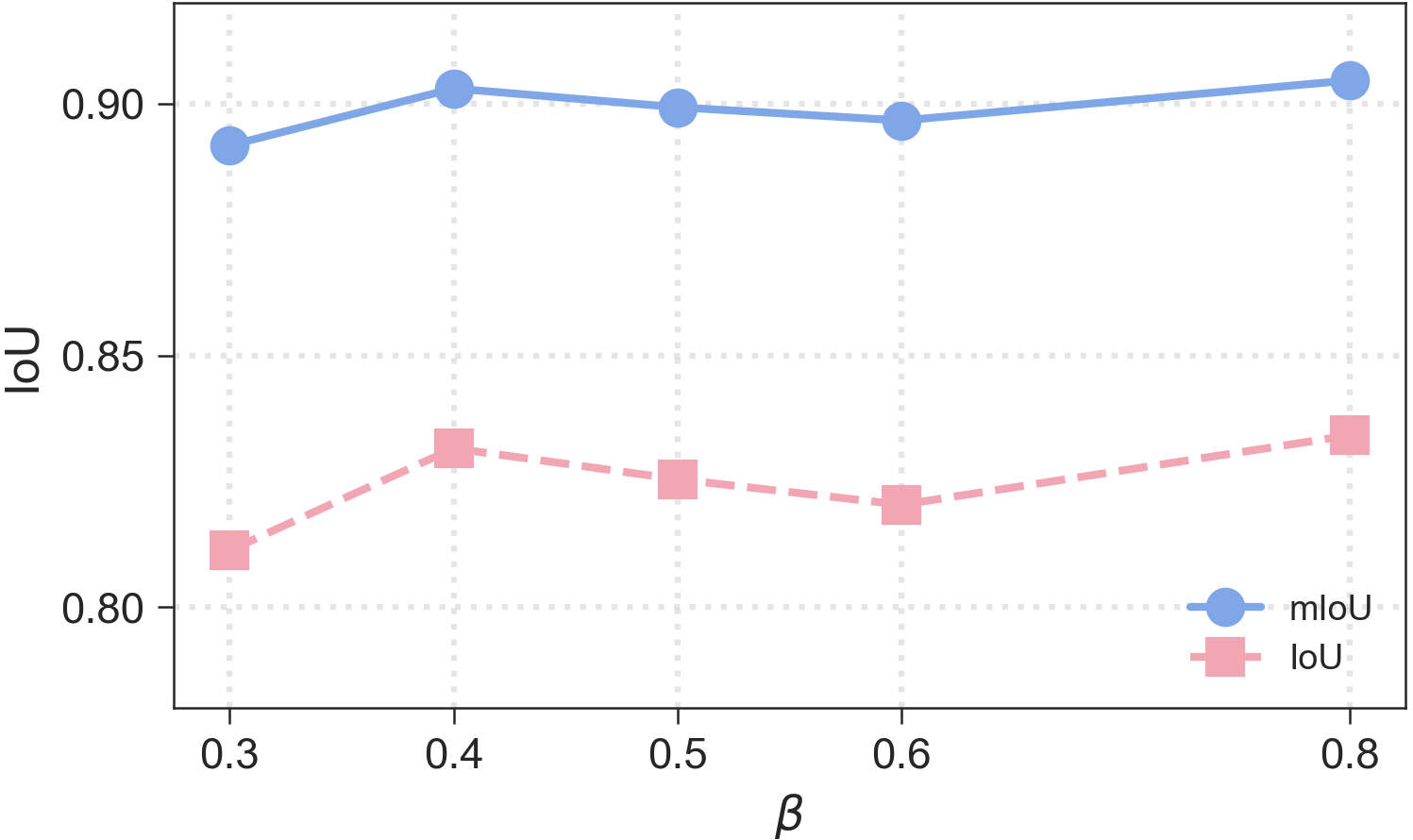}
    \caption{The effect of bias factor $\beta$ on Transformer (semantic) and CNN (spatial) feature fusion}
    \label{fig:bias}
\end{figure} 
In our Fusion mechanism, the bias factor $\beta$  (Eq. 11) control the stability and reliability of the attention-based fusion between Transformer (semantic) and CNN (spatial) feature streams.  We systematically tuned $\beta$ within the range 0.2 to 0.8 and observed that higher bias values consistently improved fusion stability and downstream accuracy (Fig. \ref{fig:bias}). In particular, $\beta = 0.8$ yielded the best results on Sen1Floods11, indicating that giving slightly more emphasis to Prithvi’s transformer-based embeddings, while still allowing the attention module to incorporate complementary CNN spatial details, produces the most robust and discriminative multi-scale representations. 

\begin{table}[ht]
\centering
\caption{Performance evaluation of Prithvi-CAFE on Sen1Flood11 compared with state-of-the-art methods. References given next to model depict source of the results. A dash (--) indicates that the value was not reported in the respective study.}
\label{flood_table}
\small 
\resizebox{\linewidth}{!}{%
\begin{tabular}{lcccc}
\hline
\textbf{Model} & \textbf{IoU W} & \textbf{mIoU} & \textbf{m-F1} & \textbf{Params (M)} \\
\hline
Prithvi 2.0 600M \cite{szwarcman2024prithvi}  & 82.50 & 90.30  & 94.80 & 650 \\
Prithvi 2.0 300M \cite{szwarcman2024prithvi}  & 82.20 & 89.70  & 97.60  & 319 \\
TerraMindv1-B    & 82.90 & \  90.60 & \   95.01 & 103 \\
MM UNet \cite{portales2025understanding}         & 73.84 & --   & --    & -- \\
DeepSARFlood \cite{sharma2025deepsarflood}            & 72.22 & --   & --    & -- \\
U-Prithvi \cite{kostejn2025u}         & 82.22 & 89.73 & --   & -- \\
RemoteCLIP \cite{marsocci2024pangaea}      & 55.18 & 72.26 & 83.83 & 83.83 \\
GFM-Swim \cite{marsocci2024pangaea}        & 52.36 & 72.60 & 82.51 & 82.51 \\
S12-DINO \cite{marsocci2024pangaea}        & 80.25 & 88.61 & 93.75 & 85 \\
Spectral GPT \cite{marsocci2024pangaea}    & 81.02 & 89.07 & --    & 600 \\
DOFA \cite{marsocci2024pangaea}            & 81.54 & 89.37 & 94.20 & 410 \\
U-Net Base \cite{li2023assessment}        & 84.03 & \textbf{90.80} & 95.40 & 31 \\
U-Net \cite{kostejn2025u}             & 80.69 & 88.84 & --    & 31 \\
DeCUR Full FT \cite{marti2025fine}        & --    & 86.87 & --    & 25 \\
Prithvi 2.0 300M LoRA \cite{marti2025fine}        & -- & 90.04 & -- & 5.5 \\
Prithvi 2.0 300M ViT Adapter \cite{marti2025fine} & -- & 88.52 & -- & 20 \\
\textbf{Prithvi-CAFE (Ours)}      & \textbf{83.41} & {90.50} & \textbf{97.80} & \textbf{45.5} \\
\hline
\multicolumn{5}{c}{\textbf{Sen1Flood11 Bolivia}} \\
\hline
Prithvi 1.0  \cite{kostejn2025u}      & 72.42 & 82.89 & --    & -- \\
U-Prithvi \cite{kostejn2025u}         & 79.68 & 87.70 & --    & -- \\
U-Net Base \cite{li2023assessment}        & 70.57 & 82.54 & --    & 31 \\
DeCUR Full FT \cite{marti2025fine}        & --    & 85.84 & --    & 25 \\
Prithvi 2.0 300M LoRA \cite{marti2025fine}        & --    & 87.57 & -- & 5.5 \\
Prithvi 2.0 300M ViT Adapter \cite{marti2025fine} & --    & 84.94 & -- & 20 \\
Prithvi 2.0 300M Full FT \cite{marti2025fine}     & --    & 82.07 & -- & 300 \\ 
\textbf{Prithvi-CAFE (Ours)}      & \textbf{81.37} & \textbf{88.87} & \textbf{96.87} & \textbf{45.5} \\
\hline
\end{tabular}
}
\end{table}

\subsection{Model’s evaluation on FloodPlanet}
The comparative evaluation on FloodPlanet \cite{zhang2025assessing} reveals strong performance by Prithvi-CAFE, achieving the highest IoU-Water (64.70), highest mIoU (68.74), and best m-F1 score (81.45) while using only 45.5M trainable parameters—significantly fewer than large foundation models such as Prithvi-2.0-600M (650M) and Prithvi-2.0-300M (319M), which obtain 61.91 and 62.03 IoU-Water, respectively (Table \ref{tab:floodplanet_k4}). TerraMind also shows competitive performance with 62.33 IoU, whereas U-Net falls short at 60.14, DOFA at 59.15, TransNorm \cite{azad2022transnorm} at 60.19, and UViT (which also uses the Prithvi encoder) at 59.16 (Table \ref{tab:floodplanet_k4}).  Prithvi-CAFE consistently outperforms these approaches by a substantial margin of 2.5–7.2 improvement in mIoU and 1.6–9 points in F1. These findings align with the Sen1Floods11 test results on a held-out test site, where Geo-Foundational Models (GFMs) demonstrated superior generalization to completely unseen locations compared to U-Net (Table \ref{flood_table}). The proposed Prithvi-CAFE model excelled in both scenarios, highlighting the combined advantage of a pretrained transformer encoder with CNN-derived features. Overall, the results highlight the effectiveness and efficiency of Prithvi-CAFE for flood segmentation under heterogeneous conditions.
\begin{table}[h!]
\centering
\caption{Comparison of state-of-the-art models on 4 fold cross validation. Mean values followed by standard deviations in parentheses.}
\label{tab:floodplanet_k4}

\resizebox{\linewidth}{!}{%
\begin{tabular}{lccc c}

\hline
\textbf{Model} & \textbf{IoU Water} & \textbf{mIoU} & \textbf{m-F1} & \textbf{Params (M)} \\
\hline
Prithvi 2.0 600M        & 61.91 (0.04) & 65.05 (0.04) & 76.07 (0.03) & 650 \\
Prithvi 2.0 300M        & 62.03 (0.02) & 66.05 (0.02) & 79.83 (0.01) & 319 \\
TerraMind               & 62.33 (0.01) & 66.19 (0.01) & 79.53 (0.01) & 103 \\
U-Net                   & 60.14 (0.01) & 64.56 (0.01) & 75.09 (0.01) & 31 \\
DOFA                    & 59.15 (0.03) & 61.52 (0.02) & 74.22 (0.02) & 116 \\
TransNorm               & 60.09 (0.04) & 64.80 (0.04) & 74.95 (0.04) & 103 \\
UViT                    & 59.16 (0.04) & 63.19 (0.04) & 72.23 (0.04) & -- \\
\textbf{Prithvi-CAFE } 
                       & \textbf{64.70 (0.02)} 
                       & \textbf{68.74 (0.03)} 
                       & \textbf{81.45 (0.02)} 
                       & \textbf{45.5} \\
\hline
\end{tabular}

}
\end{table}

\begin{figure}
    \centering
    \includegraphics[width=1\linewidth]{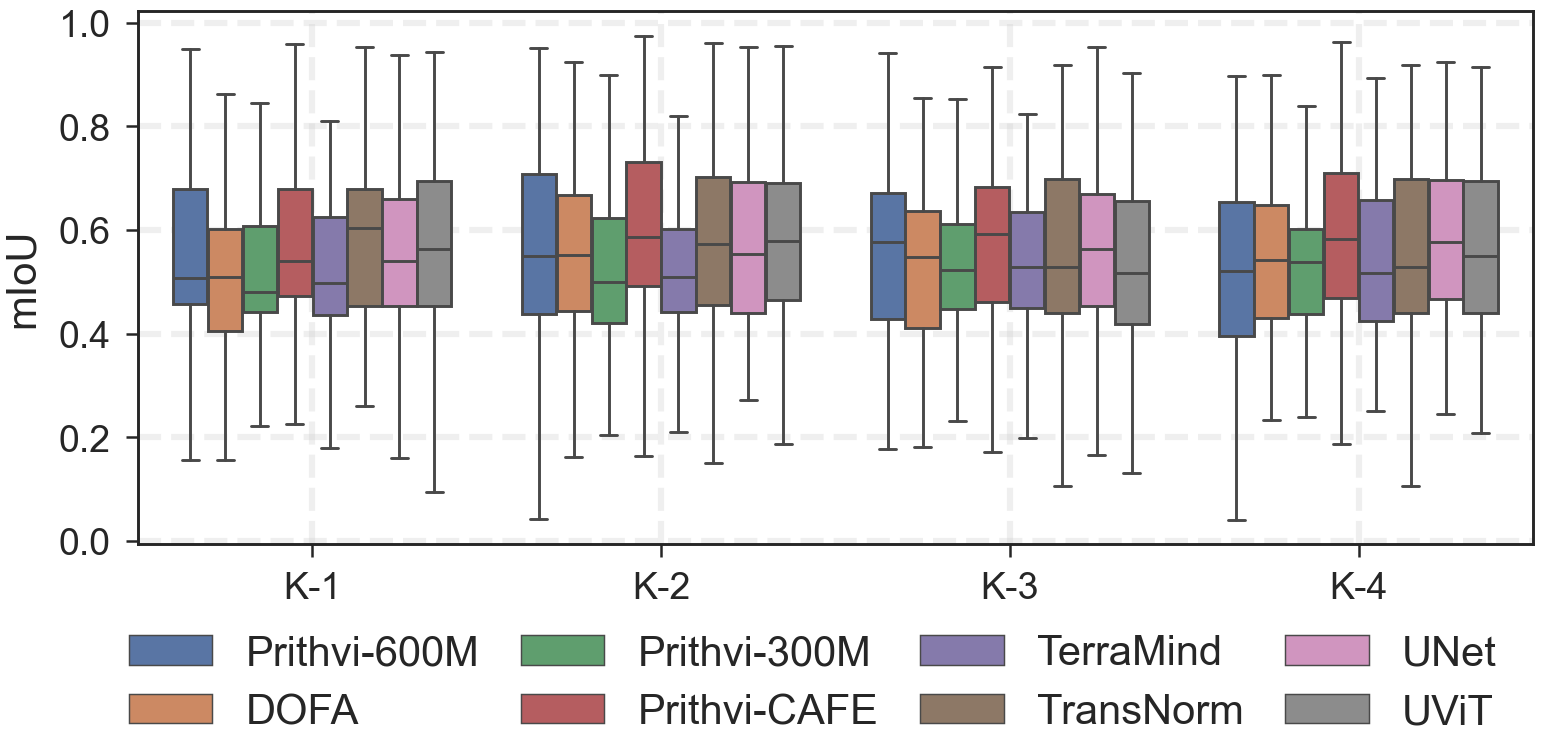}
    \caption{Box plot showing distribution of mIoU per image in four fold cross validation on FLoodPlanet}
    \label{fig:k4}
\end{figure}

The box plot showing distribution of mIoU per image across each split demonstrates consistent performance by Prithvi-CAFE, with a standard deviation of 0.02, highlighting the model’s strong generalization capability under heterogeneous conditions (Fig.\ref{fig:k4}). In contrast, other models such as TransNorm exhibit a relatively higher standard deviation of 0.04; as the figure indicates, TransNorm performs well on splits 1 and 2 but shows significantly lower performance on splits 3 and 4 (Fig. \ref{fig:k4}). Other large foundation models, such as Prithvi-2.0-600M and TerraMind-Base, have standard deviations of 0.03 and 0.14, respectively, TerraMind-Base and U-Net report the lowest standard deviation (0.01) among all.
\begin{figure}[h]
    \centering
    \includegraphics[width=1\linewidth]{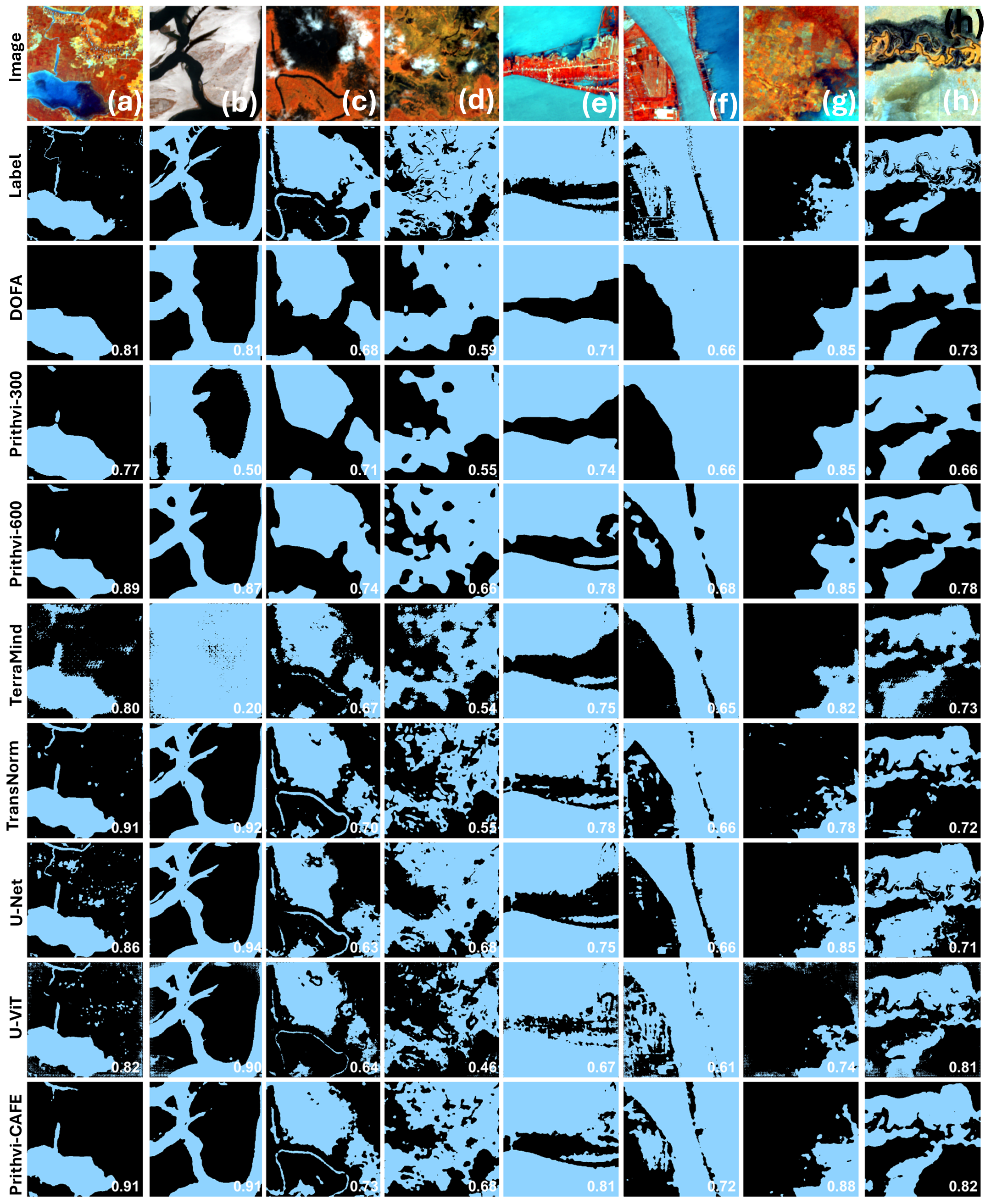}
    \caption{Comparative visual analysis of the proposed Prithvi-CAFE against other models using FloodPlanet data. Numbers in the lower-right corner indicate mIoU.}
    \label{fig:flood}
\end{figure}

The qualitative analysis also confirms the consistent outperformance of Prithvi-CAFE compared to other models. Fig. \ref{fig:flood} depicts several examples where Prithvi-CAFE performs better than the original Prithvi GFM, U-Net and TerraMind. The most interesting observations (Fig. \ref{fig:flood} (c), (f), and (h)) highlight cases where other GFMs such as Prithvi-2.0-600M, TerraMind, and DOFA capture only the overall flooding pattern, while U-Net captures fine local details. In contrast, our proposed Prithvi-CAFE successfully leverages both long-range and detailed local information, outperforming traditional CNNs and large GFMs. These observations emphasize the effectiveness of our approach in fusing long-range information from the transformer encoder with enhanced local spatial details extracted by parallel CNN blocks.
\begin{figure}[h]
    \centering
    \includegraphics[width=1\linewidth]{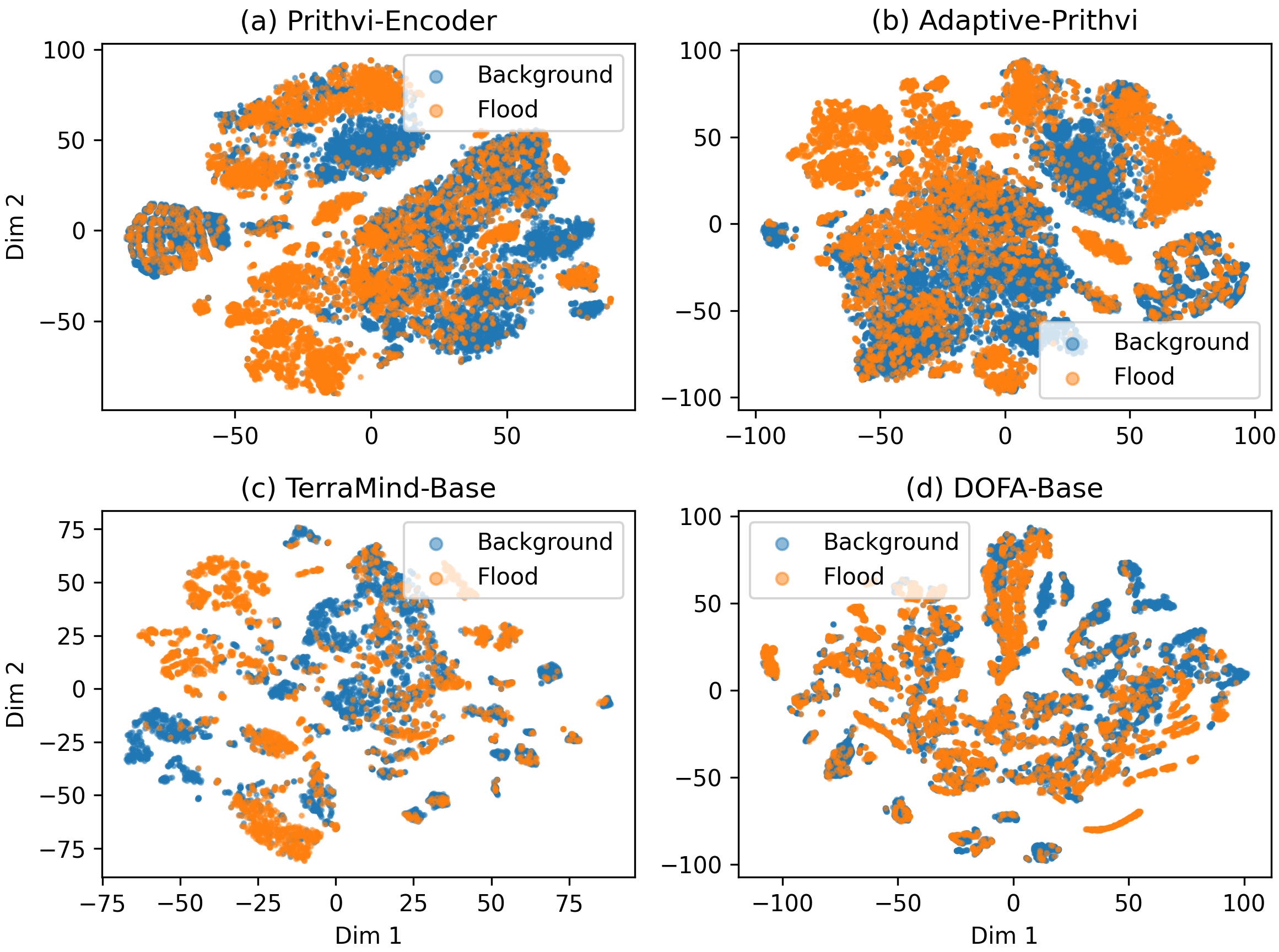}
    \caption{Visualization of feature embedding of FloodPlanet data using t-SNE plots. (a) fully fine-tune Prithvi 2.0 encoder (600M), (b) Adaptive Prithvi fine-tuning (only 45.5M parameters are trained) (c) fully fine-tune TerraMind-Base encoder, (e) fully fine-tune DOFA-Base encoder.}
    \label{fig:embeddings}
\end{figure}
In addition to the final segmentation results, we also visualize the feature embeddings generated by different encoders, including Prithvi-2.0-600M, DOFA, TerraMind, and our Adaptive Prithvi (Fig. \ref{fig:embeddings}). The visualization shows that using the efficient and fast fine-tuning method of Adaptive Prithvi, differentiation between the two classes becomes clearly visible while fine-tuning only ~7 of the total encoder parameters. Although these embeddings are not the final segmentation masks, they provide a reasonable indication of how the encoder distinguishes between classes. These embeddings are subsequently passed to the decoder to generate the final segmentation mask.
\begin{table}[h]

\centering
\caption{Ablation study on FloodPlanet dataset: effect of different modules of Prithvi-CAFE segmentation performance}
\label{table:ablataion_block}
\small
\resizebox{\linewidth}{!}{%
\begin{tabular}{@{}cc|cc|c|ll@{}}
\toprule
\multicolumn{2}{c|}{Prithvi} & \multicolumn{2}{c|}{CNN Module} & \multirow{2}{*}{M${^2}$FAF} & \multirow{2}{*}{mIoU} & \multirow{2}{*}{mDice} \\ \cmidrule(r){1-4}
w/o Adapter & Adapted & Res Block & CAM &   &       &       \\ \cmidrule(l){1-7} 
\checkmark           &         & \checkmark        & \checkmark   & \checkmark & 67.06 & 79.65 \\
            & \checkmark       &           &     &   & 65.43 & 76.81 \\
            & \checkmark       & \checkmark         &     &   & 66.83 & 78.53 \\
            & \checkmark      & \checkmark         & \checkmark   &   & 66.91 & 78.84 \\
            & \checkmark       & \checkmark         & \checkmark   & \checkmark & \textbf{68.74} & \textbf{81.45} \\ \bottomrule
\end{tabular}%
}
\end{table}
\subsection{Ablation Study}
Table \ref{table:ablataion_block}  summarizes the ablation study conducted on FloodPlanet dataset to evaluate the impact of different modules within our proposed Prithvi-CAFE segmentation framework. The first row shows the baseline configuration without Prithvi adaptation but including the CNN enhancement and M${^2}$FAF modules. This setup achieves moderate performance (67.06 mIoU, 79.65 mDice), highlighting the benefits of convolutional refinement and feature fusion, even without adapter tuning. When only the Prithvi adaptation is enabled (second row), performance decreases, indicating that adaptation alone is insufficient without complementary feature enhancement. Adding the residual block or CAM individually (third and fourth rows) gradually improves the results, showing that each component contributes meaningful spatial–channel refinement. The final configuration, which integrates all modules Prithvi adaptation, residual block, CAM, and M${^2}$FAF, achieves the highest performance. 
\begin{table}[h]
\centering
\caption{Ablation study of CNN Module Channel Configuration}
\small
\label{tab:ablation_config}
\resizebox{0.6\linewidth}{!}{%
\begin{tabular}{@{}lll@{}}
\toprule
Configuration               & mIoU  & mDice \\
\midrule
{[}32, 64, 128,   256{]}    & 65.03 & 76.45 \\
{[}64, 128,   256, 512{]}   & 66.31 & 78.12 \\
{[}128, 256,   512, 1024{]} & \textbf{68.74} & \textbf{81.45} \\ \bottomrule
\end{tabular}%
}
\end{table}
We further conducted an ablation study to investigate the impact of different CNN configurations within the CNN module of Prithvi-CAFE on segmentation performance (Table \ref{tab:ablation_config}).
The results show a clear trend: smaller channel configurations underperform due to insufficient feature representation capacity, while intermediate configurations improve performance but do not fully exploit the model’s potential. The highest configuration ([128, 256, 512, 1024]) achieves the best performance, matching the optimal full potential of Prithvi-CAFE.

\subsection{Modes of failure}
Overall, the proposed Prithvi-CAFE performed well across heterogeneous testing sites; however, the model still shows some limitations, particularly in cases of dense cloud cover (Fig. S1). Fig. S1 (see Supplementary) explicitly illustrates several cloudy scenarios and misclassifications by all models. Even under these challenging conditions, Prithvi-CAFE performs relatively better than other models. These observations highlight the limitations of large GFMs in capturing detailed local information, with significant misclassification occurring in cloud-shadow regions. In contrast, Prithvi-CAFE efficiently incorporates long-range dependencies and critical local details, giving it an edge over other models. It is noteworthy that FloodPlanet labels were originally derived from high-resolution PlanetScope imagery (3m), indicating that our model can capture very fine details where other models struggle. To further improve Prithvi-CAFE, we suggest incorporating SAR data as input, since the model can handle any number of channels. The complementary information provided by SAR is expected to enhance flood mapping capabilities under cloudy conditions.

\section{Conclusion}
Most GFMs struggle to outperform U-Net in the flood segmentation task using the benchmark Sen1Floods11 dataset, highlighting their limitations in capturing critical spatial nuances. To address these challenges, we propose Prithvi-CAFE, which leverages the Prithvi pretrained encoder, a fast and efficient fine-tuning method using adapters, and a dual-path architecture: the six bands on which Prithvi was originally pretrained are passed to the Transformer block, while the remaining channels are processed through a parallel CNN block (Residual + CAM). A multi-scale, multi-stage attention-based fusion mechanism then combines long-range information from the Transformer with detailed local information from the CNN block. The proposed Prithvi-CAFE outperforms SoTA results on Sen1Floods11, with significant improvement (6.63 mIoU) observed on geographically held-out test sites. Results on FloodPlanet further confirm the consistent superior performance of Prithvi-CAFE (2-7 mIoU) compared to other GFMs and U-Net. Visual analysis demonstrates that Prithvi-CAFE efficiently retains both long-range and local information, capturing fine details where other GFMs struggle. Our detailed observations also highlight dense cloud cover as a limiting factor for model performance.
{
    \small
    \bibliographystyle{ieeenat_fullname}
    \bibliography{ref}
}

\cleardoublepage
\newpage
\onecolumn
\appendix
\section{Appendix}
\addcontentsline{toc}{section}{Supplementary Material} 
\renewcommand{\thesection}{\Alph{section}} 
\renewcommand{\thetable}{\thesection.\arabic{table}} 
\renewcommand{\thefigure}{\thesection.\arabic{figure}} 

This example illustrates the comparative evaluation of models in a dense cloudy scene. Our observations highlight Prithvi-CAFE’s relatively better performance, even under challenging cloudy conditions.
\begin{figure}[h]
    \centering
    \includegraphics[width=0.59\linewidth]{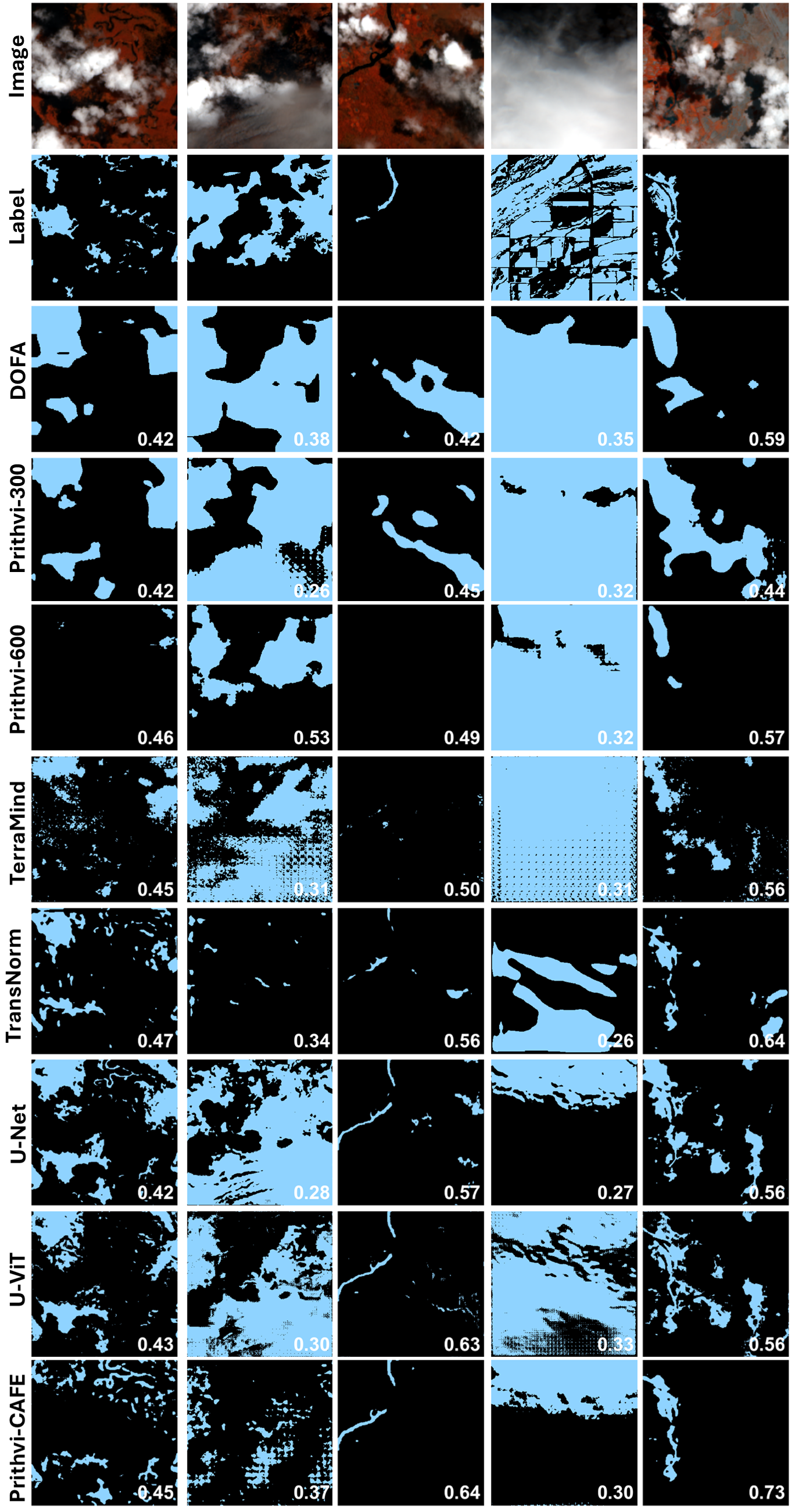}
    \caption{Figure illustrate limitation of Prithvi-CAFE in case of dense cloud cover.}
    \label{fig:suppl}
\end{figure}

\end{document}